\documentclass{article}

\PassOptionsToPackage{numbers}{natbib}
\usepackage[final]{iai_neurips_2024}

\usepackage[utf8]{inputenc} 
\usepackage[T1]{fontenc}    
\usepackage[colorlinks]{hyperref}
\usepackage{url}            
\usepackage{booktabs}       
\usepackage{amsfonts}       
\usepackage{nicefrac}       
\usepackage{microtype}      
\usepackage[dvipsnames]{xcolor}     
\usepackage{amsmath}
\usepackage{todonotes}
\usepackage{physics}
\usepackage{graphicx}
\usepackage{subcaption}
\usepackage{multirow}
\usepackage{tabularx}
\usepackage{array}
\usepackage{import}
\usepackage{tikz}
\usepackage{verbatim}
\usepackage[capitalize,noabbrev]{cleveref}
\usepackage{float}
\newcolumntype{Y}{>{\centering\arraybackslash}X}
\usepackage{wrapfig}
\usepackage{tikzscale}
\title{ProtoS-ViT: \\ Visual foundation models for sparse self-explainable classifications}

\author{
  Hugues Turb\'e$^{1}$\thanks{Corresponding author: hugues.turbe@unige.ch}\\
  \vspace{-5mm}
  \And
  Mina Bjelogrlic$^{1}$ \\
  \vspace{-5mm}
  \And
  Gianmarco Mengaldo$^{2}$ \\
  \vspace{-5mm}
  \And
  Christian Lovis$^{1}$ \vspace{5mm} \\
  $^{1}$ Division of Medical Information Sciences, Geneva University Hospitals and \\ Department of Radiology and Medical Informatics, University of Geneva, Switzerland \vspace{2mm} \\ 
  $^{2}$ Department of Mechanical Engineering, College of Design and Engineering, \\National University of Singapore, Singapore\\
}

\begin{document}

\maketitle

\begin{abstract}
Prototypical networks aim to build intrinsically explainable models based on the linear summation of concepts. Concepts are coherent entities that we, as humans, can recognize and associate with a certain object or entity. However, important challenges remain in the fair evaluation of explanation quality provided by these models. This work first proposes an extensive set of quantitative and qualitative metrics which allow to identify drawbacks in current prototypical networks. It then introduces a novel architecture which provides compact explanations, outperforming current prototypical models in terms of explanation quality. Overall, the proposed architecture demonstrates how frozen pre-trained ViT backbones can be effectively turned into prototypical models for both general and domain-specific tasks, in our case biomedical image classifiers. Code is available at \url{https://github.com/hturbe/protosvit}.
\end{abstract}

\section{Introduction}

As deep learning (DL) models are increasingly used for decision making, transparency is becoming a critical issue. 
Lack of transparency has been repeatedly identified as a key barrier for adoption of DL models in high-risk areas, including the healthcare sector~\cite{topol2019high}. In this sense, research around explainable AI (XAI) has seen the development of a number of methods which can be broadly separated into two areas: i) post-hoc interpretability methods, and ii) self-explainable models. Post-hoc interpretability methods are applied on trained models and typically provide a relevance or saliency map that reveals the importance of each input feature to a certain output ~\cite{samek2021explaining}.  This work instead focuses on models which are explainable by design, or self-explainable model (SEM), bypassing the need for post-hoc interpretability. Part-prototype models are special SEM aimed at learning concepts that can be linearly combined to classify images \cite{chen2019looks}.

Along the development of XAI models, the evaluation of the explanations provided by these models is critical. Several works showed that while post-hoc interpretability methods have some attractive properties: they are model agnostic (e.g.~SHAP~\cite{lundberg2017unified}), and they do not affect the performance of the underlying DL model, they also suffer from some critical drawbacks. Key drawbacks include a lack of faithfulness to explaining the model~\cite{sanity_adebayo,reliable_dylan, turbe2023evaluation} and sensitivity to negligible perturbations~\cite{ghorbani2019interpretation, dombrowski2019explanations}. More recently, explanations provided by prototypical networks have also been shown to be inaccurate, mainly because they often do not correctly localize important parts of the image for the classification \cite{sacha2023interpretability,carmichael2024pixel} as well as not representing coherent concepts in the input space \cite{nauta2023pip,hoffmann2021looks}. 

A comprehensive set of 12 properties (Co-12) to evaluate explanation quality has been defined in \cite{nauta_anecdotal_2023}.  Evaluating model's interpretability has been shown to be difficult as it often relies on human's apriori knowledge \cite{sanity_adebayo} or might be plagued by distribution's shift induced by the interpretability evaluation methods \cite{hooker2019benchmark}. The FunnyBirds framework \cite{hesse2023funnybirds} was designed to evaluate several aspects of explanation quality while addressing the difficulties listed above. It includes a synthetic dataset along a number of metrics. While not designed specifically for SEM models it covers several aspects found in the Co-12 properties. In addition, the dataset can be used to adapt previous metrics designed specifically to evaluate prototypical part models. Based on these aspects, the contributions of this work are the following:
\begin{enumerate}
\item We provide an extensive set of quantitative metrics and qualitative evaluation adapted to prototypical-part models and tackling different issues identified in the literature.  Applying these metrics on state-of-the-art (SOTA) part-prototypical model,  highlights important issues regarding the correctness and contrastivity of the explanations obtained with these models.

\item We propose a novel architecture, ProtoS-ViT, addressing the shortcomings identified with previous SEM models. ProtoS-ViT, leverages frozen foundation models (ViT) as the backbone to provide SOTA performance in terms of explanation i) \textit{correctness}, ii) \textit{compactness}, using no more than seven prototypes for the benchmark datasets which cover both general and biomedical tasks; iii) \textit{consistency}, explanations are consistent, semantically and visually coherent; iv) \textit{contrastivity}, explanation correctly identifies discriminative parts of the image while being competitive in terms of \textit{classification performance} on a wide range of datasets. ProtoS-ViT is \textit{computationally efficient} as it only requires training a lightweight head.
\end{enumerate}

\section{Related Work}\label{sec:related}

Research on outcome explainability using SEMs that are explainable by design has been very active in the last few years. Many self-explainable classifiers are based on the prototypical part architecture following the ProtoPNet model~\cite{chen2019looks}. Prototypical part models aim to extract concepts that can be linearly combined to classify images. While part-projection has been commonly used to align prototypes with specific patches in the training data, recent studies  \cite{nauta2023pip,ma2024looks} have moved away from this approach. We argue that part-projection conflicts with neuroscience theories of human brain concept learning mechanisms, which propose two models for concept representation: (i) the exemplar model, where concepts are represented by multiple exemplars, and (ii) the prototype model, where concepts are abstracted from specific exemplars \cite{zeithamova2019brain}. Forcing prototypes to match specific patches fails to align with either the exemplar or prototype model of concept representation in human cognition.

Starting from ProtoPNet work~\cite{chen2019looks} different methods were developed to improve the classification performance, the faithfulness of the explanations as well as reduce the number of prototypes used by the model to make a decision~\cite{rymarczyk2021protopshare, donnelly2022deformable, wang2023learning, nauta2023pip}. 
These improvements were mainly achieved by devising new ways to create the prototypes and introducing new losses to lower the semantic gap between the prototypes and meaningful concepts from images. The developed architectures  use different variations of CNN backbones including VGG~\cite{simonyan2015very}, ResNet~\cite{he2016deep} and DenseNet~\cite{huang2017densely}, followed by a linear classifier. Other approaches replaced the final linear classifier with a decision tree. For instance, ProtoTree combined a CNN backbone with a decision tree~\cite{Nauta_2021_CVPR}, while the ViT-NeT architecture combines a vision transformer (ViT) backbone with a neural tree decoder~\cite{pmlr-v162-kim22g}.

Explainability is multifaceted, imposing a number of desiderata for a model to be explainable \cite{nauta_anecdotal_2023}. The Co-12 properties \cite{nauta_anecdotal_2023} aim to define 12 properties that comprehensively evaluate the quality of an explanation. We quickly introduce the most relevant properties from~\cite{nauta_anecdotal_2023} along the stability property from~\cite{huang2023evaluation}:
\begin{enumerate}
    \item \textbf{Correctness}: Whether the explanation faithfully represents the model's behavior.
    \item \textbf{Completeness}: How much of the model's behavior is captured by the explanation.
    \item \textbf{Consistency}: Whether similar inputs have similar explanations \cite{nauta_anecdotal_2023}, with its extension for prototypical part networks to include the prototype consistency in the input space \cite{huang2023evaluation}.
    \item \textbf{Contrastivity}: Whether the explanation correctly captures parts of the image that are discriminant for the predicted class.
    \item  \textbf{Compactness}: Whether the explanation is compact.
    \item \textbf{Composition}: The explanation presentation should reflect the model's behavior.
    \item \textbf{Stability}: Prototype attribution should be stable under small perturbations, such that perturbations invisible to the human's eyes do not change the prototype attribution \cite{huang2023evaluation}. 
\end{enumerate}

We identify two important flaws in how current XAI evaluations are performed, which we address in this work: i) lack of precise part annotations, and ii) human apriori bias.  First, several research evaluate consistency~\cite{huang2023evaluation, carmichael2024pixel, nauta2023pip} or stability~\cite{huang2023evaluation} on the center location of the object parts provided in the CUB dataset, with a box of arbitrary size often drawn around the centre to see if a prototype corresponds to a given object part. Instead, in this work, we leverage the precise part annotations provided in the FunnyBirds dataset to avoid this issue. Regarding apriori human bias, the evaluation of an explanation contrastivity has often been based again on CUB~\cite{wang2023learning,nauta2023pip}, evaluating whether the prototypes used for classification lie over the bird or the background. This introduces a human bias in the evaluation of the model interpretability as the environment of a bird might be used by the model in classifying bird species.

One common concern for SEM models and more specifically prototypical part networks is the spatial misalignment of the explanations: ``Here does not correspond to there''~\cite{carmichael2024pixel}. Given that the models have a receptive field that can reach $100\%$ of the initial image, there is no assurance that the embedding of a patch is directly correlated to the same position in the input image. More evidence of the latter issue has also recently been raised by several authors -- see e.g.,\cite{sacha2023interpretability,carmichael2024pixel}. However, none of the metrics found in the litterature directly evaluate spatial alignment. Indeed the metrics presented in~\cite{carmichael2024pixel} such as \textit{ROT} evaluate the correctness of the model and not directly the spatial alignment. This distinction is shown to be important in our discussion. The metric proposed by \cite{sacha2023interpretability} is based on adversarial noise added to the input pixels outside the pixels activated by the prototype with the largest activation in an image. Given this augmentation, we argue that this method essentially evaluates a model's robustness to adversarial attacks (\textit{stability}) and not the spatial alignment. A  model robust to such attacks could perform well without necessarily encoding local information. We therefore observe that currently, no single metric can alone guarantee the spatial alignment of the proposed explanations.

%
%
\section{Methodology}\label{sec:methods}
\subsection{Benchmark for evaluation of prototypical-part models}
Based on the properties presented above, metrics from the FunnyBirds framework are used to evaluate the correctness, completeness, consistency, and contrastivity of the explanations. We refer the reader to the paper that introduces the framework~\cite{hesse2023funnybirds} for more details on the metrics used for the evaluation of these properties. The part importance function \textit{PI} used to compute the metrics is adapted to prototypical part models as described in Appendix~\ref{appendix:funny_birds}. This 
adaptation follows a recent work on how to design the importance function for prototypical part models~\cite{oplatek2024revisiting}. Consistency and stability properties are evaluated by adapting the corresponding metrics developed by~\cite{huang2023evaluation} to the FunnyBirds dataset, leveraging the precise part-annotations available for this dataset. More details on the adaptation can be found in Appendix~\ref{app:interpretability_metrics}. This change allows for a finer evaluation of the explanations by considering the full similarity map and not only the top relevance. Explanation compactness is evaluated following the metrics defined in~\cite{nauta2023pip}, that is, the global size to measure the total number of prototypes retained by the model to make its predictions across the whole task; and the local size, to measure the average number of prototypes used to make a prediction on a single image. We restrict the number of local prototypes to the ones used for the predicted class following the definitions in~\cite{nauta2023pip}. 

\subsection{ProtoS-ViT architecture}
\begin{figure}[ht]
    \centering
    \resizebox{.95\linewidth}{!}{\includegraphics{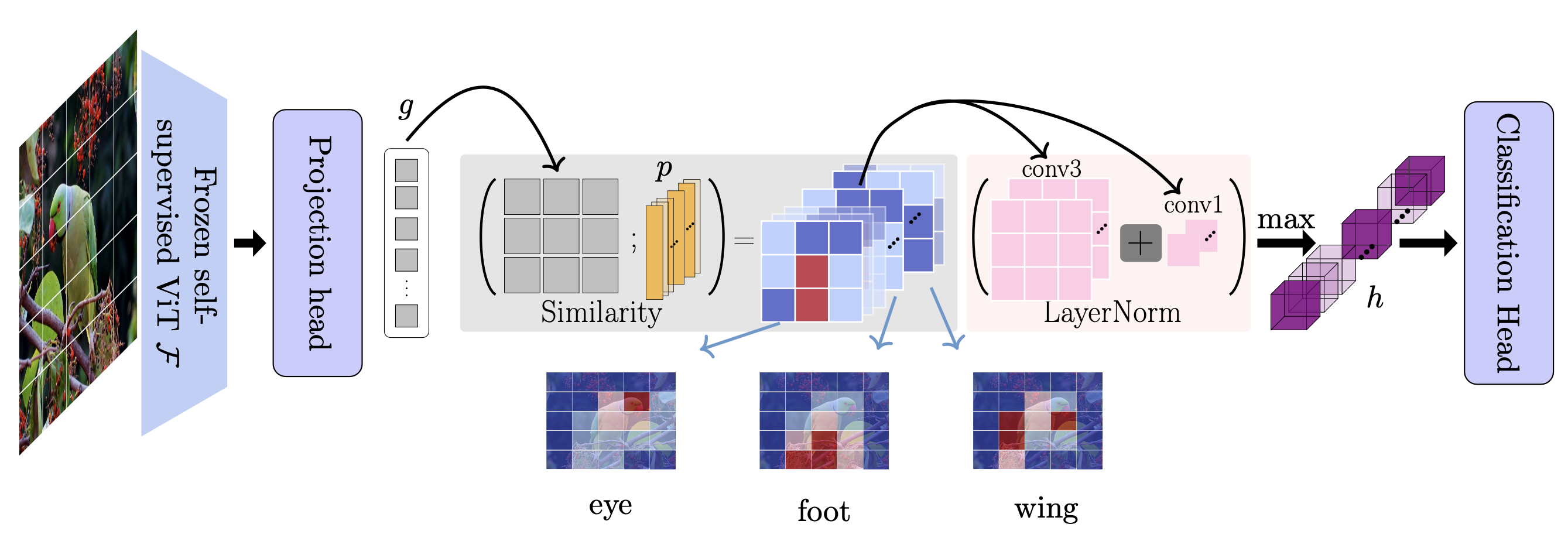}}
    \vspace{-5pt}
    \caption{Model architecture. 
    The grey box depicts the similarity head. 
    The pink box indicates the operations forming the prototypical head. 
    Transparency of the elements aims to reflect the model's sparsity. 
    Bottom: similarity maps interpolated from the similarity head.}
    \label{fig:model_schema}
\end{figure}

An overview of the global architecture is depicted in~\cref{fig:model_schema} and is described in more detail next.
Consider a classification task that consists in mapping an image $x \in \mathbb{R}^{H \times W \times C}$ to a labelled target $y\in \mathbb{N}^K$ where $H,W,C$ represent, respectively, the height, width and number of channels of the input image, and $K$ is the number of classes. 
The input image is fed to a pre-trained feature extractor $\mathcal{F}: x \rightarrow f_i\in \mathbb{R}^{C_e}$ for patch index $i \in [1, \cdots, I]$, and $I = \frac{H}{s}\cdot \frac{W}{s}$, with $s$ indicating the patch size of the encoder, and $C_e$ the size of the image embedding dimension.

Following the pre-trained feature extractor, a projection head map consisting of three convolution layers with residual connections maps the features to the corresponding projected features $g_i \in \mathbb{R}^{D}$. 
This organization is inspired by ~\cite{seong2023leveraging} for unsupervised segmentation. 
All convolutions at this stage have a $1\times1$ kernel size to retain local information. 
The projected features are then compared to the prototypes by calculating their cosine similarity: 
\begin{equation}
    S_{i,j} = \cos <g_i, p_j >.
\end{equation} 
where $p_j$ denotes prototype $j$ in the set of all prototypes $\mathcal{P}=\{p_j \in \mathbb{R}^D\}$, with $j \in [1, \cdots, J]$ where $J$ represents the initial number of learned prototypes and $D$ their dimension. 
The prototypes similarity distribution for each patch is then normalised with a softmax function so that the normalised similarity is equal to $\tilde{S}_{i,j} = \sigma_i \left (S_{i,j} / \tau \right)$. 

Once the prototype similarity distribution is known for each patch, the second step aims to determine the importance of the prototype distribution at the image level towards the final class with a novel prototypical head. 
Most prototypical models only use the maximal value for each prototype across the image as an input to the final classification head \cite{chen2019looks, gautam2023looks, gautam2022protovae} such that the prototype score $h_j = \max_i \tilde{S}_{i,j}$. 
However, this operation prevents the model from learning how the distribution of a prototype presence across the image influences its importance. 
To tackle this issue, we introduced depthwise convolutions with independent kernels for each prototype. Independent kernels are key for the score to properly reflect the importance of a single prototype presence with no interactions between prototypes. 
In addition, to model the presence of the prototype at different scales, two convolutions were introduced following insights from the \textit{Inception} architecture \cite{szegedy2015going}; a convolution with a kernel of size $1 \times 1$ and another one with size $3\times3$. 
The output of the two convolutions applied to the matrix $\tilde{S}_{i,j}$ is then summed and normalised by a LayerNorm:
\begin{equation}
    h_j = \max \left \{\textit{LayerNorm} \left( \textit{Conv}_{1\times 1} (\tilde{S_j}) + \textit{Conv}_{3\times 3} (\tilde{S_j}) \right) \right \}.
\end{equation}
The max operator is finally applied to the sum and values of $h_j$ below 0.1 are set to 0 when doing inference. 
The final classification head is then a simple linear classifier with weights $W$ restricted to being positive to improve the explainability of the model. 
This linear layer takes as input the vector $h$ that indicates the global score of each prototype in the input image. 
The linear layer converts this score into a class based on the importance of each prototype towards the class of interest. 
For the rest of the work, we define the importance matrix $\mathbf{I}=\left(i_{k,j}\right) \in \mathbb{R}^{K\times J}$ with the importance ${i}_{k,j}$ of prototype $j$ toward class $k$ as: 
\begin{equation}
    i_{k,j}= W_{k,j} \times h_j ,
\end{equation}
It is important to note that concepts are not specific to a single class, allowing the model to share common concepts across classes and reducing the overall number of prototypes required to perform a given classification task. This is particularly relevant for tasks where some classes might share many concepts in common, as shown in the experiment section.

In order to fulfil the compactness properties described in the introduction, the model should provide a classification using as few concepts as required for a single image (local size of the explanation), as well as using the smallest number of coherent concepts for the entire task (global size) to avoid redundancy of the learned prototypes. In our work,  compactness is promoted with a regularization loss applied on the importance matrix $\mathbf{I}$, namely the Hoyer-Square (HS)~\cite{yang2019deephoyer}:
\begin{equation}
     \mathcal{L}_{HS} = \alpha \frac{\abs{\mathbf{I}}^2}{\norm{\mathbf{I}}_2^2} + \gamma \Vert \mathbf{I} \Vert _2.
     \label{eq:sparsity_loss_app}
\end{equation}
In order to minimise the number of concepts used for each prediction, we set $\alpha=\gamma=0.01$. 
In addition to these terms, we also adopt the tanh-loss $\mathcal{L}_T$ devised by \cite{nauta2023pip}: 
\begin{equation}
    \mathcal{L}_T = -\frac{1}{J} \sum_{j}^J \log  ( \tanh{(\sum_i^ {I\times B} \tilde{S}_{i,j}) + \epsilon} )
\end{equation} 
where $B$ is the batch size. 
This last loss is key for the model not to collapse under the pressure of the sparsity loss $\mathcal{L}_{HS}$ at the beginning of the training procedure, enforcing that each prototype is at least present once in each batch. 
The total loss function is therefore:
\begin{equation}
    \mathcal{L} = \mathcal{L}_{CE} + \phi \mathcal{L}_{HS} +  \mathcal{L}_T
\end{equation}
where $\mathcal{L}_{CE}$ is the cross-entropy loss between the model's prediction and the target, and $\phi$ the sparsity loss factor. Nomenclature can be found in Appendix~\ref{app:nomenclature} 

%
%
\section{Experiments}\label{sec:experiments}
\begin{table}[h]
\caption{Accuracy (Acc.), Global Size (Glob. Size), and Local Size (Loc. Size) for different models on the general datasets. \textbf{Bold} indicates the best score for the given metric. $^\star$Additional evaluation of this architecture reported a lower accuracy of 84.51\%~\cite{xue2022protopformer}. }
\label{table:summary_results}
\begin{tabularx}{\textwidth}{c|YYY|YYY|YYY|YYY}
\hline
                  & \multicolumn{3}{c}{CUB}       & \multicolumn{3}{c}{CARS}      & \multicolumn{3}{c}{PETS}      & \multicolumn{3}{c}{Funny Birds} \\ 
Method            & Acc. $\uparrow$ & Glob. \newline Size $\downarrow$  &  Loc. \newline Size $\downarrow$ & Acc. $\uparrow$ & Glob. \newline Size $\downarrow$ &  Loc. \newline Size $\downarrow$ & Acc. $\uparrow$ & Glob. \newline Size $\downarrow$  &  Loc. \newline Size $\downarrow$ & Acc. $\uparrow$ & Glob. \newline Size $\downarrow$   &  Loc. \newline Size $\downarrow$ \\ \midrule
DINO-L/14       & 90.5 &    NA        &    NA       & 90.1 &     NA       &    NA       & 96.6 &      NA      &   NA        &       &             &           \\
ProtoPNet         & 79.2 & 2000       &           & 86.1 & 1960       &           &      &            &           & 94    & 500         &           \\
ProtoTree         & 82.2 & 202        &           & 86.6 & 195        &           &      &            &           &       &             &           \\
ProtoPShare       & 74.7 & 400        &           & 86.4 & 480        &           &      &            &           &       &             &           \\
ProtoPool         & 85.5 & 202        &           & 88.9 & 195        &           &      &            &           &       &             &           \\
PIP-Net           & 84.3 & 495        & \textbf{4}         & 88.2 & 515        & \textbf{4}         & 92   & 172        & \textbf{2 }        & 81.2  & 47          & \textbf{1}         \\
ViT-NeT           & \textbf{91.6$^\star$} &            &           & \textbf{93.6} &            &           &      &            &           &       &             &           \\
PixPNet           & 81.8 &       2000     & 10        &      &            &           &      &            &           &       &             &           \\
ST-ProtoPNet      & 86.1 & 8000       &           & 92.7 & 3920       &           &      &            &           & \textbf{99.6}  & 1000        & 20        \\
ProtoS-ViT (ours) & 85.2 & \textbf{39}         & 6         & 93.5 & \textbf{54}         & 7         & \textbf{95.2} & \textbf{44}         & 4         & 96.8  & \textbf{26}          & 6        
\end{tabularx}
\end{table}

\textbf{Backbones} We choose  DINOv2~\cite{oquab2023dinov2} and OpenClip~\cite{software_openclip} as the backbone for general tasks. Both these models have demonstrated strong performance across a range of computer vision benchmarks. DINOv2 is particularly interesting as it has been used for unsupervised segmentation tasks achieving SOTA performance and demonstrating the quality of local information obtained with this model~\cite{oquab2023dinov2, kim2023causal}.
To further demonstrate the versatility of the proposed approach we also apply our model to three Biomedical tasks using the ViT from BioMedCLIP~\cite{zhang2023biomedclip}. Full experimental setups and datasets are described in respectively  Appendix~\ref{app:experimental} and Appendix~\ref{appendix:Dataset}.

\textbf{Baselines} We compare the proposed approach to a non-explainable baseline (DINOv2 ViT-L/14, with a linear classifier reporting results from the initial model publication \cite{oquab2023dinov2}) along SOTA explainable prototypical models, namely ProtoPNet~\cite{chen2019looks}, ProtoTree~\cite{Nauta_2021_CVPR}, ProtoPShare~\cite{rymarczyk2021protopshare}, ProtoPool~\cite{protopool}, PIP-Net~\cite{nauta2023pip}, ViT-Net~\cite{pmlr-v162-kim22g}, ST-ProtoPNet~\cite{wang2023learning}, PixPNet~\cite{carmichael2024pixel}. We present all results found in the litterature, that is either in the paper presenting the model or in further work. In addition, we also benchmark our architecture along PIP-Net and ST-ProtoPNet on the FunnyBirds dataset \cite{hesse2023funnybirds}. These models were selected because ST-ProtoPNet consistently achieves the highest accuracy among prototypical models in the literature, while PIP-Net offers the most compact explanations in terms of both local and global size. We additionally retrained the PIP-Net architecture on FunnyBirds with a DINOv2 ViT-B/14 to differentiate the impact of the backbone architecture from the overall prototypical part architecture. Results are shown in Appendix~\ref{app:backbone} along with additional results to analyse the impact of training or freezing the backbone.
\begin{wrapfigure}{r}{0.5\textwidth} 
    \centering
    \includegraphics[width=.5\textwidth]{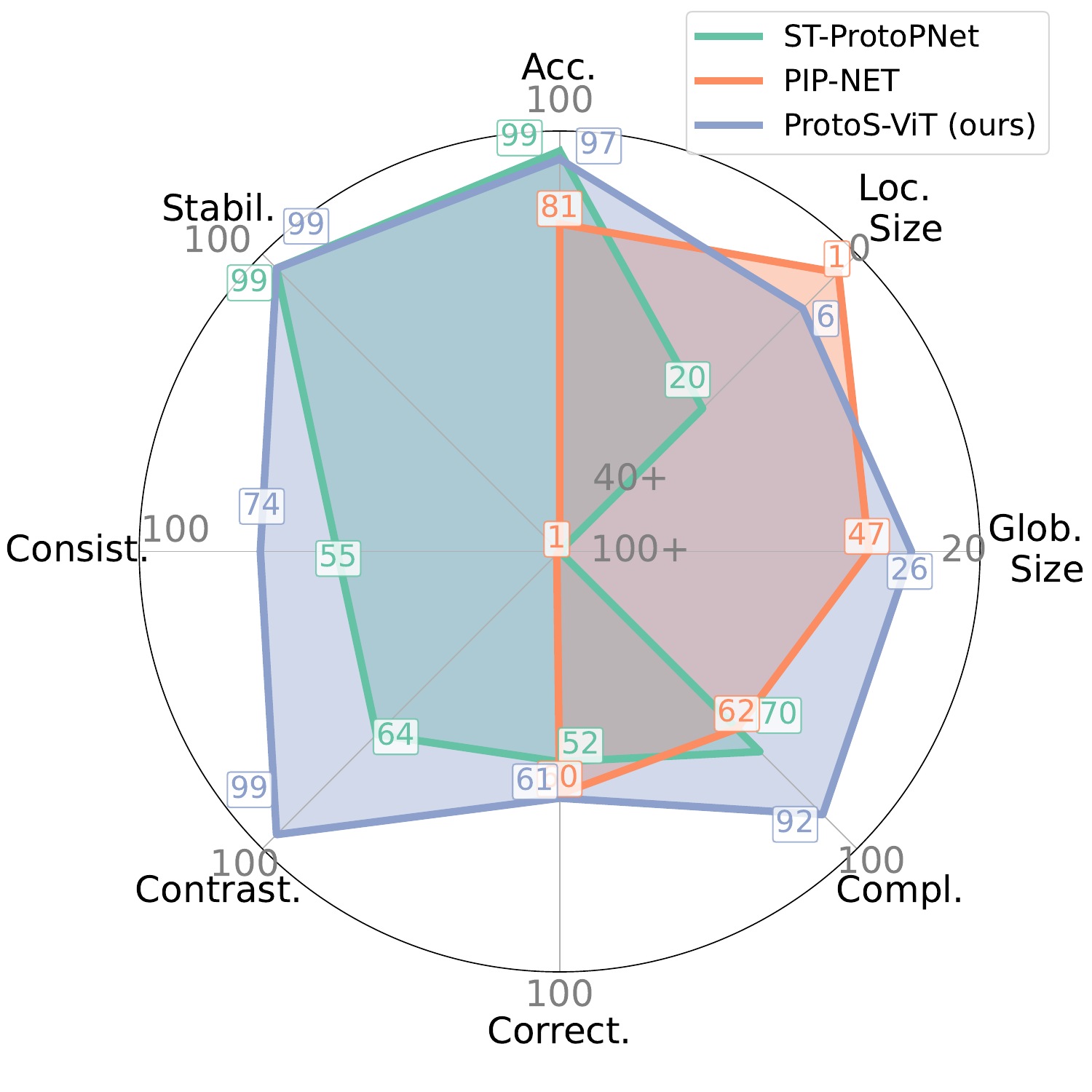}
    \caption{Radar plot summarizing model performance both in terms of Accuracy (Acc.) as well as explainability quality with the following metrics Global Size (Glob. Size), and Local Size (Loc. Size), Completeness (Compl.), Correctness (Correct.), and Contrastivity (Contrast.), Consistency (Consist.), and Stability (Stabil.).}
    \label{fig:radar_plot}
\end{wrapfigure}
Model performance in terms of accuracy as well as local and global size are shown for the general datasets in Table~\ref{table:summary_results}. Results for the Biomedical datasets are shown in Appendix~\ref{app:biomed}. Figure~\ref{fig:radar_plot} shows a radar plot of six explainability properties described above along with the classification accuracy for the proposed model and two SOTA prototypical models, namely PIP-Net and ST-ProtoPNet. This quantitative assessment of the quality of the explainability was complemented by a user-study on FunnyBirds with the results presented in Appendix~\ref{app:User-Studies}.
The accuracy for our model on the CUB dataset is an average over four runs, where the standard deviation was respectively 0.14, 0.11 and 2.9 for the accuracy, local size and global size. Results for the general dataset with the OpenCLIP backbone are presented in Appendix~\ref{app:openclip_results}.

\begin{figure}[ht]
\begin{center}
\centerline{\includegraphics[width=.95\columnwidth]{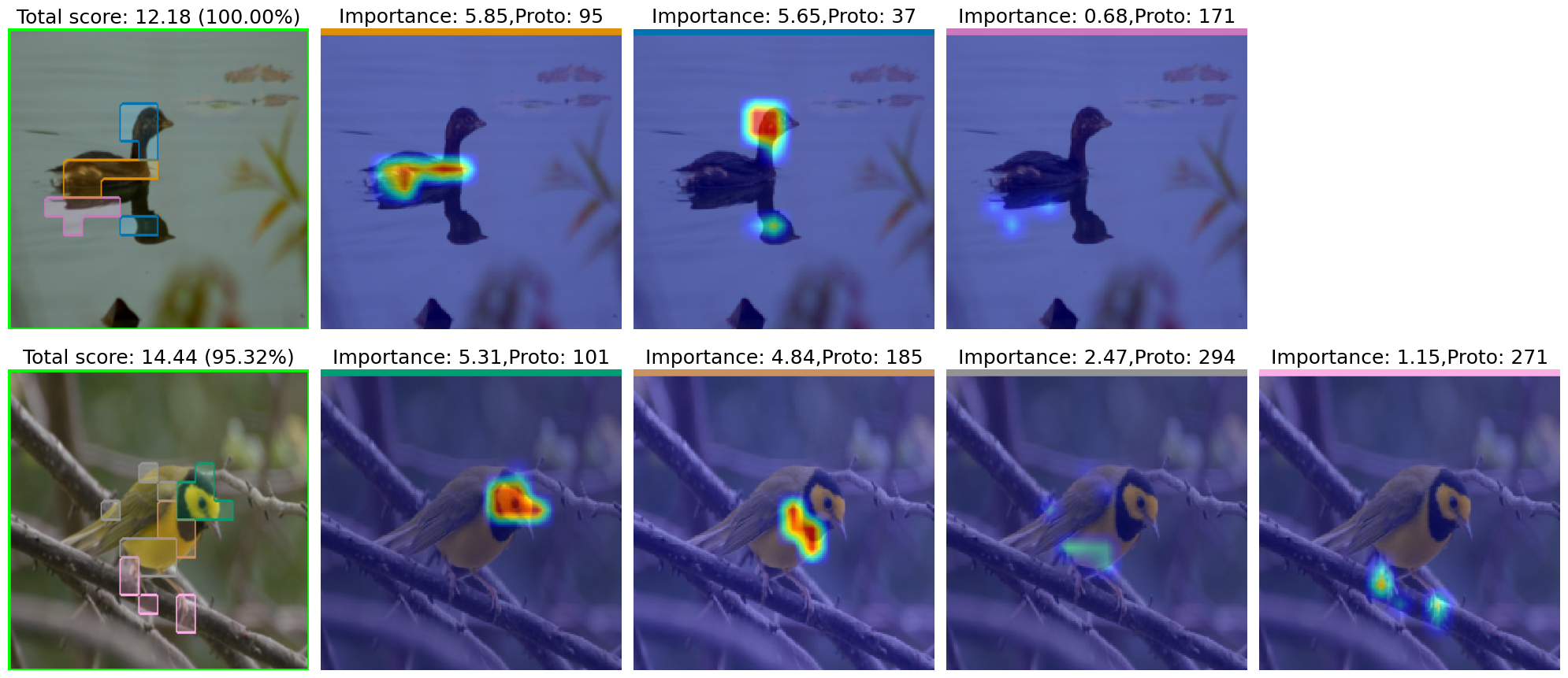}}
\caption{Score sheet for predictions on two random samples of the CUB dataset. Each row shows a prediction on a different sample. The first column indicates the position of the top four prototypes. Each subsequent column shows a prototype along with its importance towards the predicted class. Above the first column, we present the total score for the predicted class as well as how much of this score is explained by the prototypes shown in the figure.  \vspace{-0.5em}}
\label{fig:score_sheet_cub}
\end{center}
\end{figure}

We show score sheets with predictions on two instances from the CUB dataset in Figure~\ref{fig:score_sheet_cub} with the relative importance of the four most important prototypes for each prediction. The first image in each row shows the location of the four prototypes while the heatmap in the subsequent images represents the similarity map between each patch following the projection head and the corresponding prototype $p_j$. Given that the backbone output has a reduced spatial dimension (e.g., DINOv2 has a stride of 14), we interpolate the similarity map back to the original input resolution. Above each prototype, we indicate the corresponding importance, and we retain a consistent color scheme across images to represent identical concepts. Above the first image of each row, we show the predicted score as well as the percentage of this score explained by the prototypes shown in the figure. Additional score sheets for all datasets are shown in Appendix~\ref{app:additional_score_sheet} and in Supplementary materials~\cite{zenodo}.
%
\section{Discussion}
\label{sec:discussion}

The aim of this work was twofold: i) to provide a comprehensive set of metrics to identify current issues in common prototypical part networks, ii) propose a novel architecture aiming to address some of these flaws.

The evaluation of explanation quality is complex and requires a multifaceted evaluation. Based on the requirements set out in the literature, the proposed set of metrics is the first to allow a thorough evaluation of prototypical part models. This global assessment is key to highlight critical issues that might not appear when only considering specific aspects of the evaluation. For instance, while PIP-Net achieves good classification accuracy and consistent explanations, as observed both visually by inspecting the prototypes and through the quantitative evaluation (see Figure~\ref{fig:radar_plot}), it fails to produce contrastive explanations as highlighted by the contrastive metric equal to zero. The patches highlighted by the prototype do not represent the discriminative portions of the images. Instead, the model encoded discriminative features found across the images in given patch embeddings with no direct relation to the features found at this precise location. 

Another issue with models found in the literature is the local and global size of the explanations. Models such as ST-ProtoPNet have up to 40 prototypes per image while showing activation maps for only a few prototypes, which breaks the completeness and composition requirements set by~\cite{nauta_anecdotal_2023}. In the score sheets presented in this work, see for example Figure~\ref{fig:score_sheet_cub}, we present the percentage of the predicted score explained by the prototypes shown in the score sheet. With a small local size, a very large fraction of the explanation can be easily shown to the user.
As shown in Table~\ref{table:summary_results}, the proposed architecture matches or exceeds the accuracy of comparable SOTA prototypical networks\footnote{ViT-Net achieves the top accuracy on CUB and CARS but do not support simple (linear) case-based reasoning on prototypes, as noted in~\cite{carmichael2024pixel}.}.  Regarding explanation quality, the radar plot in Figure~\ref{fig:radar_plot} shows that ProtoS-ViT achieves the highest overall explainability score. ST-ProtoPNet performs well in explanation faithfulness, with similarity maps accurately reflecting pixel importance for classification, as seen in the evaluation of contrastivity and correctness. However, its lower completeness indicates that parts of the image outside the explanations still influence predictions. ProtoS-ViT performs strongly in completeness and contrastivity, capturing all relevant pixels. Further evaluation shows that the model is robust to part deletion, compensating by enhancing the contribution of remaining parts; see Appendix~\ref{appendix:funny_birds}. An ablation study demonstrated that the design of the prototypical head is key to reduce the global size of the developed model; see Appendix~\ref{app:ablation}. A trade-off involving the global size of the model must be carefully considered, as prototypical models need to generate a diverse array of prototypes to accurately classify images. However, these models often produce redundant prototypes, which leads to an increased global size and hinders the explainability of their results. Going beyond quantitative metrics, user studies were used to evaluate this trade-off, demonstrating that the developed architectures effectively reuse concepts across different classes.; see Appendix~\ref{app:correlation_matrix} as well as producing consistent prototypes from a human perspective; see Appendix~\ref{app:User-Studies}.

A key aspect of our architecture was to retain the backbone frozen throughout the training.
ViT models like DINOv2 have been shown to produce semantically consistent embeddings with local information~\cite{darcet2023vision}. By maintaining the backbone frozen along a projection head with a receptive field equal to one, we were able to retain this spatial alignment offered by foundational ViT models and obtain explanations quality surpassing other models. We conducted a study to demonstrate that training the backbone along the rest of the architecture might indeed improve classification performance but at the cost of lower explanation quality, see Appendix~\ref{app:backbone}. Although spatial alignment is not directly measurable through a single metric, it is a key aspect to obtain good explanations with prototypical part networks (as demonstrated in our analysis of PIP-Net). Further work should therefore focus on how the spatial alignment of backbones such as ViT can be retained while training the model end-to-end for specific tasks. In addition, experiments on biomedical datasets showed classification performance on par with non-explainable baseline. Further works will need to investigate in more details the possible applications of this architecture for biomedical datasets.

\begin{ack}
HT, MB and CL acknowledge financial support from the \textit{Fondation Carlos et Elsie De Reuter}. GM acknowledges support from
MOE Tier 1 grant no. 22-4900-A0001-0: "Discipline-Informed Neural Networks for Interpretable
Time-Series Discovery". In addition, all authors thank Julien Ehrsam, MD for his insightful comments
related to the evaluation of the models on biomedical tasks. The computations were performed at University of Geneva using Baobab HPC service.
\end{ack}

\bibliographystyle{unsrtnat}
\bibliography{main}

\appendix

\section*{Appendix}

\section{Nomenclature}
\label{app:nomenclature}
\begin{table}[H]
\centering
\caption{Notations and symbols used in this paper.}
\begin{tabular}{@{}lll@{}}
\toprule
                      & Symbol              & \multicolumn{1}{c}{Definition} \\ \midrule
Variables             & $x \in \mathbb{R}^{H \times W \times C}$        & Sample (image)                         \\
                      & $y\in \mathbb{N}^K$  & Labeled target               \\
                      & $K$                & Number of classes                     \\
                      & $H$                  & Sample height          \\
                      & $W$                  & Sample width  \\
                      & $C$                  & Sample number of channels (3 for RGB images) \\
                      & $s$            & Patch size of encoder           \\
                      & $i \in [1, \cdots, I]$          & patch index \\
                      & $I = \frac{H}{s}\cdot \frac{W}{s}$          & Number of patches \\
                       & $C_e$                  &  Patch embedding dimension \\ 
                      & $J$                   & Initial number of prototypes \\
                      & $\mathcal{P}=\{p_j \in \mathbb{R}^D\}$        & Learnable set of prototypes            \\
                      & $p_{j}$           & Prototypes $j$  \\
                      & $j \in [1, \cdots, J]$          & Prototype index \\
                     & $D$                  &  Prototype dimension \\
                      & $g_{i}$           & Projected sample feature $i$  \\
                      & $\mathbf{I}=\left(i_{k,j}\right) \in \mathbb{R}^{K\times J}$        & Importance matrix            \\
                      &$h_j$ & Prototype score \\        
                       \midrule
Operators             & $\mathcal{F}: x \rightarrow f_i\in \mathbb{R}^{ C_e}$         & Encoder operator      \\
                      & $S_{i,j}$   & Cosine similarity between projected sample $g_i$ and prototype $p_j$      \\ 
                      & $\tilde{S}_{i,j}$   & Normalised cosine similarity     \\ 
                      \bottomrule
\end{tabular}
\label{table:variable_summary}
\end{table}

\section{Experimental Setup}
\label{app:experimental}

The proposed architecture is implemented in PyTorch~\cite{paszke2019pytorch}. 
First, all images were resized to a pixel resolution of $224\time224$ using random resizing and cropping during training and center cropping at test time. 
Image augmentation was also performed during training using the AugMix method~\cite{hendrycks2019augmix}. 
The large version of DINOv2, ViT-L/14 with registers~\cite{oquab2023dinov2,darcet2023vision} as well as OpenCLIP ViT-L/14 \cite{software_openclip} were tested as backbone for the general classification tasks. 
For biomedical tasks, the ViT encoder from BioMedCLIP~\cite{zhang2023biomedclip} was used as the backbone. 
The backbone was frozen and the rest of the architecture trained for 80 epochs, with 10 epochs used for the warm-up. 
The learning rate (lr) increased linearly during the warm-up to a value of 0.01 with subsequent application of cosine decay. 
In addition, each model was initialized with 300 prototypes $\mathrm{P} = \{p_j\}_{j=1}^{300}$ with $p_j \in \mathbb{R}^{512}$. 

All models were trained on an internal cluster with each model trained on a single NVIDIA GeForce RTX 3090, 12 cores and 64 GB of memory. All models are trained for 80 epochs with an AdamW optimiser and a base learning rate equal to 0.01. The learning is progressively increased for 15 warm-up epochs and then progressively following a cosine-decay schedule. $\phi$ and $\rho$ were both set equal to 1. With this configuration, individual models were trained in between one and three hours. 

\subsection{Setup for baseline models}
ST-ProtoPNet and PIP-Net were trained on the FunnyBirds in order to act as a baseline across the set of metrics presented in this work. Both models were trained following the baseline parameters found in the corresponding article that introduces the respective models. Parameters for ST-ProPNet are found in Table~\ref{tab:parameters_ST-ProtoPNet} and parameters for PIP-Net are listed in Table~\ref{tab:parameters_PIP-NEt}.

\begin{table}[H]
    \centering
    \caption{Hyperparameters for ST-ProtoPNet }
    \begin{tabular}{ll}
        \toprule
        Parameter & Value \\
        \midrule
        Backbone model & Densenet 161 \\
        Image size & $224\times224$ \\
        Batch size & 80 \\
       Protototype shape & $(1000,64, 1, 1)$ \\
        Prototype activation function & log \\
        LR joint optimizer & \{features:  1e-4, add on layers: 3e-3, prototype vectors: 3e-3\} \\
        LR joint step size & 10 \\
        Warm LR & \{add on layers: 3e-3,  prototype vector: 3e-3\} \\
        LR last layer & 1e-4 \\
        Epochs train & 20 \\
        Warmup epochs & 10 \\
        Push start & 100 \\
        Push epochs & [100,110,120]\\
        
        \bottomrule
    \end{tabular}
    
    \label{tab:parameters_ST-ProtoPNet}
\end{table}

\begin{table}[h]
    \centering
    \caption{Hyperparameters for PIP-Net}
    \begin{tabular}{ll}
    \toprule
       
        Parameter & Value \\
        \midrule
        Backbone Model & Convnext Tiny 26 \\
        Batch Size & 64 \\
        Batch Size Pretrain & 128 \\
        Epochs & 60 \\
        Optimizer & Adam \\
        Learning Rate & 0.05 \\
        Learning Rate Block & 0.0005 \\
        Learning Rate Network & 0.0005 \\
        Weight Decay & 0.0 \\
        Number of Features & 0 \\
        Image Size & 224 \\
        Freeze Epochs & 10 \\
        Epochs Pretrain & 10 \\
        \bottomrule
    \end{tabular}
    \label{tab:parameters_PIP-NEt}
\end{table}

\section{Dataset description}
\label{appendix:Dataset}
The datsets used in the study are either general purpose datasets (CUB-200-2011, referred as CUB, Stanford Cars, referred as CARS, and Oxford-IIIT Pets referred as PETS), medical datasets (ISIC 2019, RSNA, and LC25000), and one synthetic dataset designed for evaluating part-prototypical models (FunnyBirds). The seven datasets details (including the licence type) are described below:\\
\\
\textbf{CUB-200-2011} \cite{wah2011caltech}: The Caltech-UCSD Birds-200-2011 dataset is a dataset containing 11,788 images across 200 bird species. Each species is represented by roughly 60 images, and the dataset includes detailed annotations such as species, bounding boxes, and part locations. The CUB-200-2011 dataset is publicly available and can be used under the Creative Commons Attribution (CC-BY) license.\\

\textbf{Stanford Cars} \cite{krause20133d}: The Stanford Cars dataset contains 16,185 images of 196 classes of cars, with each class typically corresponding to a make, model, and year of a specific car. The dataset includes annotations for the car model, bounding boxes, and viewpoints. The Stanford Cars dataset licence is unknown. \\


\textbf{Oxford-IIIT Pets} \cite{parkhi2012cats}: The Oxford-IIIT Pet dataset consists of 7,349 images of 37 different breeds of cats and dogs. Each image includes a class label, species, and detailed pixel-level segmentation annotations. The dataset is available under the Creative Commons Attribution-NonCommercial-ShareAlike (CC BY-NC-SA 4.0) license.\\


\textbf{ISIC 2019} \cite{tschandl2018ham10000,codella2018skin,combalia2019bcn20000}: The ISIC 2019 dataset contains 25,331 dermoscopic images representing nine different types of skin lesions, with associated ground truth diagnoses. The dataset is part of the International Skin Imaging Collaboration (ISIC) and is available for research purposes under the CC BY-NC 4.0 license. \\

\textbf{RSNA Pneumonia Detection} \cite{shih2019augmenting}: The RSNA Pneumonia Detection Challenge dataset includes 30,000 annotated chest X-ray images, with labels indicating the presence or absence of pneumonia. This dataset was created for the RSNA 2018 Machine Learning Challenge and is freely available for non-commercial use under the terms provided by the RSNA, typically aligning with the CC BY-NC-SA 4.0 license.\\

\textbf{LC25000 (Lungs)} \cite{borkowski2019lung}: The LC25000 dataset includes 25,000 histopathology images of lung tissue, categorized into three classes: lung adenocarcinoma, lung squamous cell carcinoma, and benign lung tissue. The dataset is openly available for research and educational purposes under a Creative Commons Attribution (CC BY 4.0) license.


\textbf{FunnyBirds} \cite{hesse2023funnybirds}: The FunnyBirds dataset consists of 50 500 images (50k train, 500 test) of synthetic 50 bird species. The authors manually designed 5 bird parts: eyes (3 types), beak (4 types), wings (6 types), legs (4 types) and tail (9 types) to construct the 50 classes. The data set is openly available under the Apache-2.0 license.

\clearpage
\section{Additional score sheets}
\label{app:additional_score_sheet}
We present score sheet for results on the CARS and PETS dataset in Figure~\ref{fig:add_score_sheet}. Additional results can be found in the Supplementary Materials~\cite{zenodo} 
\begin{figure}[H]
\begin{center}
 \begin{subfigure}[b]{0.9\textwidth}
\includegraphics[width=\textwidth]{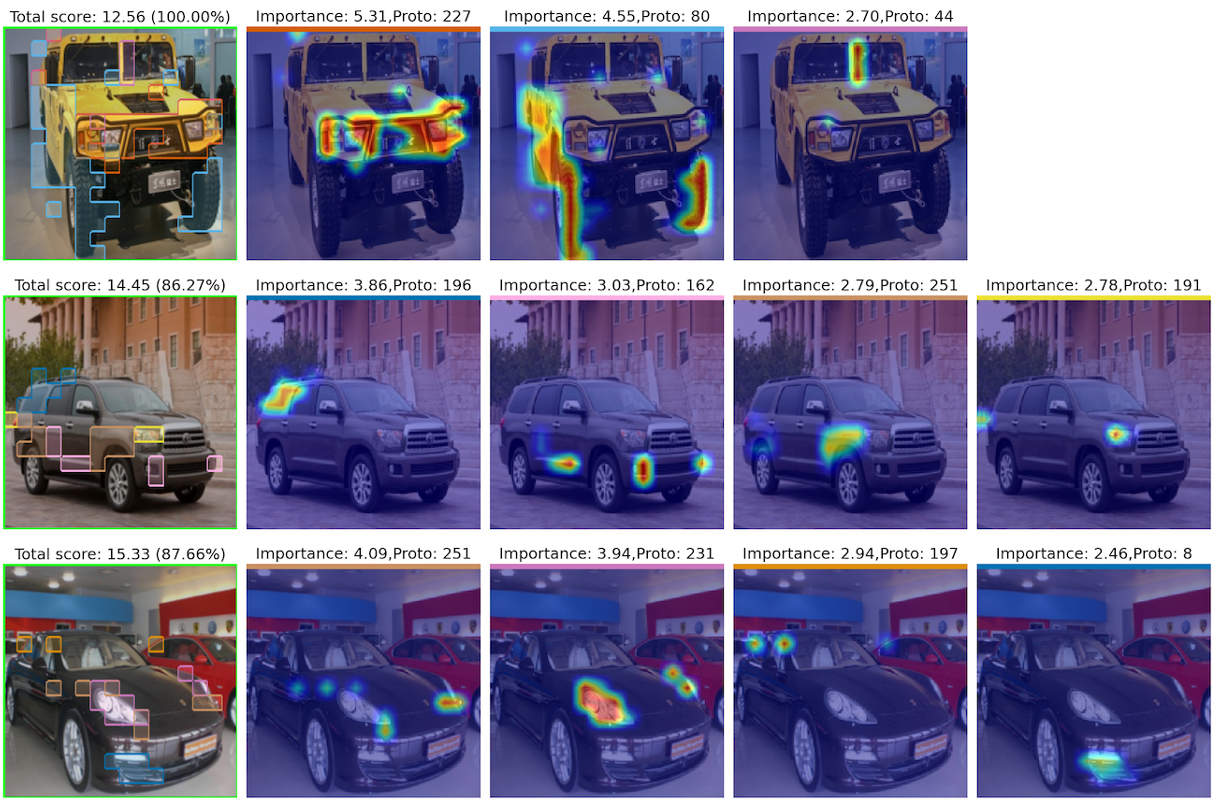}
\caption{}
\label{fig:score_sheet_cars}
\end{subfigure}
 \begin{subfigure}[b]{0.9\textwidth}
\includegraphics[width=\textwidth]{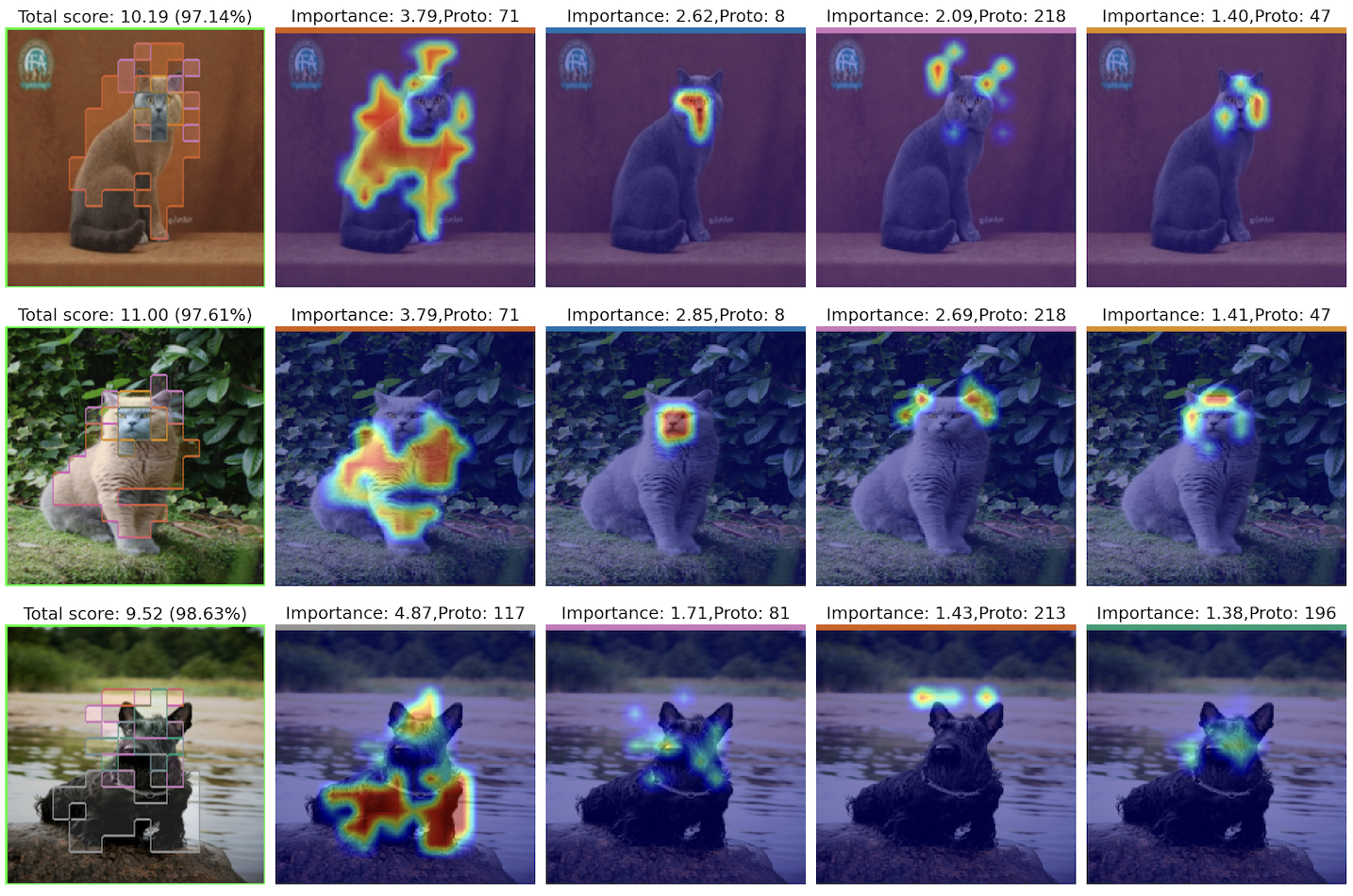}
\caption{}
\label{fig:score_sheet_pets}
\end{subfigure}
\caption{Score sheet for predictions on three random samples of the CARS (a) and PETS (b) dataset. Each row shows a prediction on a different sample. The first column indicates the position of the top four prototypes. Each subsequent column shows a prototype along with its importance towards the predicted class. Above the first column, we present the total score for the predicted class as well as how much of this score is explained by the prototypes shown in the figure.}
\label{fig:add_score_sheet}
\end{center}
\end{figure}

\section{Results with OpenCLIP backbone}
\label{app:openclip_results}
In order to evaluate the influence of the backbone on the results, the proposed architecture was evaluated with on the general datasets with OpenCLIP-L as its backbone. Results for this study are presented in Table~\ref{tab:general_openclip}. Interestingly, in the table above, the model outperforms our baseline configuration only on the CARS dataset which is the only dataset where the OpenCLIP model evaluated using a simple classification head outperforms the DINOv2 model~\cite{oquab2023dinov2}.
\begin{table}[H]
\centering
        \caption{Accuracy (Acc.), Global (Glob.) Size, and Local (Loc.) Size comparison of different models on general datasets with OpenCLIP-Large backbone}
        \label{tab:general_openclip}
        \begin{tabular}{lccc}
        \toprule
          & Acc. $\uparrow$ & Glob. Size $\downarrow$ & Loc. Size  $\downarrow$  \\ \midrule
        CUB  & 79.1 & 27 & 5 \\
        CARS & \textbf{93.8} & \textbf{50} & 6 \\
        PETS & 93.8 & 37 & 4 \\
        \bottomrule
    \end{tabular}
\end{table}
\section{Results on Biomedical datasets}
\label{app:biomed}
The developed architecture was further tested with the ViT from BioMedCLIP~\cite{zhang2023biomedclip} in order to demonstrate the usefulness of the proposed methods on specialised tasks such as biomedical tasks. Results for three tasks are presented in Table~\ref{table:biomedical_results}.
\begin{table}[H]
\centering

\caption{Accuracy (Acc.), Global (Glob.) Size, and Local (Loc.) Size comparison on three biomedical classification tasks. $^\dagger$ BiomedCLIP was evaluated on zero shot classification on LUNGS and 100-shot on RSNA. In addition accuracy for both models are extracted from graphs in the corresponding model publication~\cite{zhang2023biomedclip}.}
\label{table:biomedical_results}
\begin{tabularx}{0.7\textwidth}{llYYY}
\toprule
& Method & Acc. $\uparrow$ & Glob. \newline Size $\downarrow$ & Loc. \newline Size $\downarrow$ \\ 
\midrule
ISIC & \textbf{ProtoS-ViT (ours)} & 77.5 & 13 & 4.5 \\ \midrule
\multirow{2}{*}{RSNA} & BiomedCLIP & \textbf{83}$^\dagger$ & \\ 
 & \textbf{ProtoS-ViT (ours)} & 82.8 & 9 & 4 \\ \midrule
 \multirow{2}{*}{LUNGS}& BiomedCLIP & $65^\dagger$  \\
 & \textbf{ProtoS-ViT (ours)} &\textbf{100} & 21 & 7 \\
\midrule
\end{tabularx}
\end{table}

\subsection{Samples from the RSNA dataset}
\begin{figure}[H]
 \begin{subfigure}[b]{0.49\textwidth}
\includegraphics[width=\textwidth]{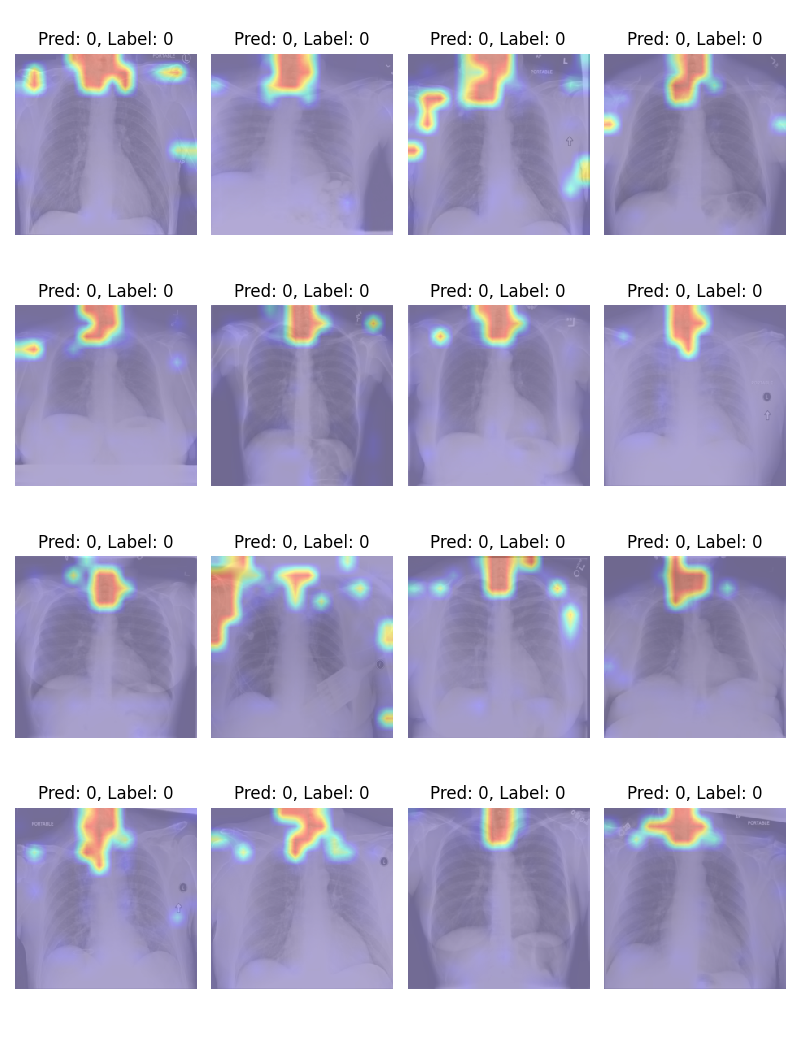}
\caption{}
\label{fig:rsna_0}
\end{subfigure}
 \begin{subfigure}[b]{0.49\textwidth}
\includegraphics[width=\textwidth]{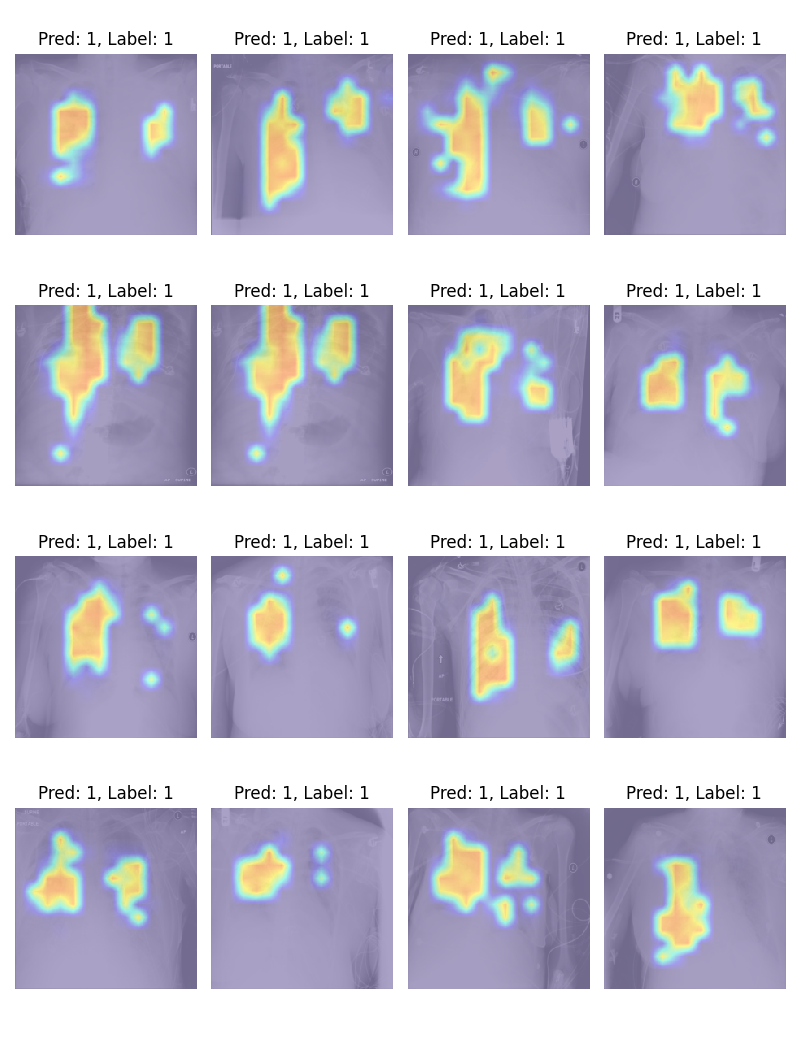}
\caption{}
\label{fig:rsna_1}
\end{subfigure}
\caption{Random sample with activation for a prototype associated with the absence (a) and presence (b) of a penuomnia}
\end{figure}

The RSNA dataset is labeled to indicate the presence or absence of pneumonia on chest X-rays. To diagnose pneumonia on chest radiographs, clinicians focus on identifying areas that show opacification of airspaces or consolidation of lung parenchyma. Interestingly, a clinician observed that prototypes associated with the presence of pneumonia consistently lay within the lungs and appeared to identify white regions corresponding to opacification or consolidation. In contrast, prototypes associated with the absence of pneumonia were located outside the lungs and seemed to lack any obvious clinical significance. One possible explanation for these irrelevant prototypes is that the model effectively learned to identify signs of pneumonia but then generated unrelated prototypes to increase the score for the absence of pneumonia, functioning similarly to a bias in the classification head. Examples of prototypes associated with both the absence and presence of pneumonia are shown in Figure~\ref{fig:rsna_0} and Figure~\ref{fig:rsna_1}, respectively.

\section{Ablation studies}
\label{app:ablation}
Ablation studies were carried to demonstrate the effectiveness of the prototypical head and sparsity loss.  For the experiment without the prototypical head, the latter was replaced with a simple max operation over the similarity matrix $\Tilde{S}$. The experiment without the sparsity loss, was conducted by replacing the loss in equation~\ref{eq:sparsity_loss_app} with an L1 norm on the weight matrix of the classification head. The results of these two experiments are presented in Table~\ref{table:ablation_study}. A study to assess the usefulness of the second kernel in the prototypical head is found in Table~\ref{table:ablation_kernel}. Finally, the impact of the weighting terms in the sparsity loss on classification accuracy, local size and global size are shown in Table~\ref{table:weights_loss}.
\begin{table}[ht]
\centering
\caption{Results from the ablation studies, presenting the baseline model compared to the architecture w/o the prototypcial head and w/o the sparsity loss. \textbf{Bold} indicates the best score for the given metric.}
\label{table:ablation_study}
\begin{tabularx}{\textwidth}{clYYlYYlYY}
\toprule
                      & \multicolumn{3}{c}{CUB}          & \multicolumn{3}{c}{Cars}         & \multicolumn{3}{c}{PETS}         \\ \cmidrule{2-10}
                      & Acc. $\uparrow$ & Glob. \newline Size $\downarrow$ & Loc. \newline Size  $\downarrow$ & Acc. $\uparrow$ & Glob. \newline Size  $\downarrow$ & Loc. \newline Size  $\downarrow$ & Acc.$\uparrow$ & Glob. \newline Size  $\downarrow$ &Loc. \newline Size  $\downarrow$ \\ \cmidrule{2-10}
Baseline              & 85.2   & \textbf{39}                &\textbf{ 6}   & \textbf{93.5}  & 54          & \textbf{7}       & 95.2  & \textbf{44 }         & \textbf{4   }       \\
w/o prototypical head   & \textbf{85.4}            & 142               & 20             & 93.5           & 148         & 19       & \textbf{95.9}              & 164         & 15          \\
w/o sparsity loss     &  84.6           &   44   &7             & 92.8          & \textbf{50}  & \textbf{7}               & 95.2            & \textbf{44 }         & \textbf{4 }  \\ \bottomrule
\end{tabularx}
\end{table}

\begin{table}[h!]
\centering
\caption{Ablation study for ProtoS-ViT with a single kernel in the prototypical head with size (1,1). }
\label{table:ablation_kernel}
\begin{tabular}{lccc}
\toprule
Dataset & Accuracy & Glob size & Loc size \\ \midrule
CUB & 85 & 37 & 7 \\
CARS & 93 & 44 & 7 \\
PETS & 95 & 56 & 6 \\
ISIC & 76 & 18 & 7 \\
RSNA & 83 & 7 & 3 \\
LUNGS & 100 & 28 & 9 \\ \bottomrule
\end{tabular}
\end{table}

\begin{table}[h!]
\centering
\caption{Study on the impact of the weighting term in the loss.}
\label{table:weights_loss}
\begin{tabular}{ccccc}
\toprule
$\alpha$ & $\gamma$ & Accuracy & Local size & Global size \\ \midrule
0.01 & 0.01 & 85.2 & 6 & 39 \\
0.1 & 0.01 & 85.2 & 4 & 34 \\
0.01 & 0.1 & 86.7 & 4 & 113 \\
0.001 & 0.01 & 86.0 & 8 & 51 \\ \bottomrule
\end{tabular}
\end{table}

\section{FunnyBirds methodology and results}
\label{appendix:funny_birds}
The FunnyBirds framework devised by Hesse et al~\cite{hesse2023funnybirds} relies on a part importance function $\text{PI}(\cdot)$ that needs to be adapted to the chosen explanation method. We adapt the $\text{PI}(\cdot)$ to reflect prototypical approaches. For each prototype $p_j$, we normalise the corresponding similarity map such that it sums to one and then multiply it by the corresponding importance $i_{j,k}$. All metrics computed on the FunnyBirds dataset to evaluate the quality of the explanation are presented for the proposed architecture, as well as for PIP-Net and ST-ProtoPNet in Table~\ref{table:funny_bird_score}.

\begin{table}[ht]
\centering
\caption{FunnyBirds evaluation metrics.}
\label{table:funny_bird_score}
\begin{tabular}{lcccc}
\toprule
                       \multicolumn{1}{c}{Metric}          & \multicolumn{1}{c}{Abbreviations}         & \multicolumn{3}{c}{Value}         \\
 &  & ST-ProtoPNet & PIP-NET & ProtoS-ViT (ours) \\  \midrule
Controlled synthetic data check & CSDC & 0.78  & 0.45  & \textbf{0.94} \\ 
Preservation check & PC               & 0.69  & 0.20   &\textbf{ 0.97}   \\
Deletion check & DC                   & 0.67   & 0.29   & \textbf{0.92} \\
Distractability & D                   & 0.69  & \textbf{0.92}  & 0.90 \\
Background independence & BI          & \textbf{1}  & \textbf{1}   & \textbf{1} \\
Single deletion & SD                  & 0.52  & 0.60  &\textbf{ 0.61} \\
Target sensitivity & TS               & 0.64  & 0.01  & \textbf{0.99} \\ \midrule
Mean explainability score & mX        & 0.62     &  0.41 & \textbf{0.84}  \\ \bottomrule
\end{tabular}
\end{table}

\begin{figure}[ht]
\begin{center}
\includegraphics[width=0.8\textwidth]{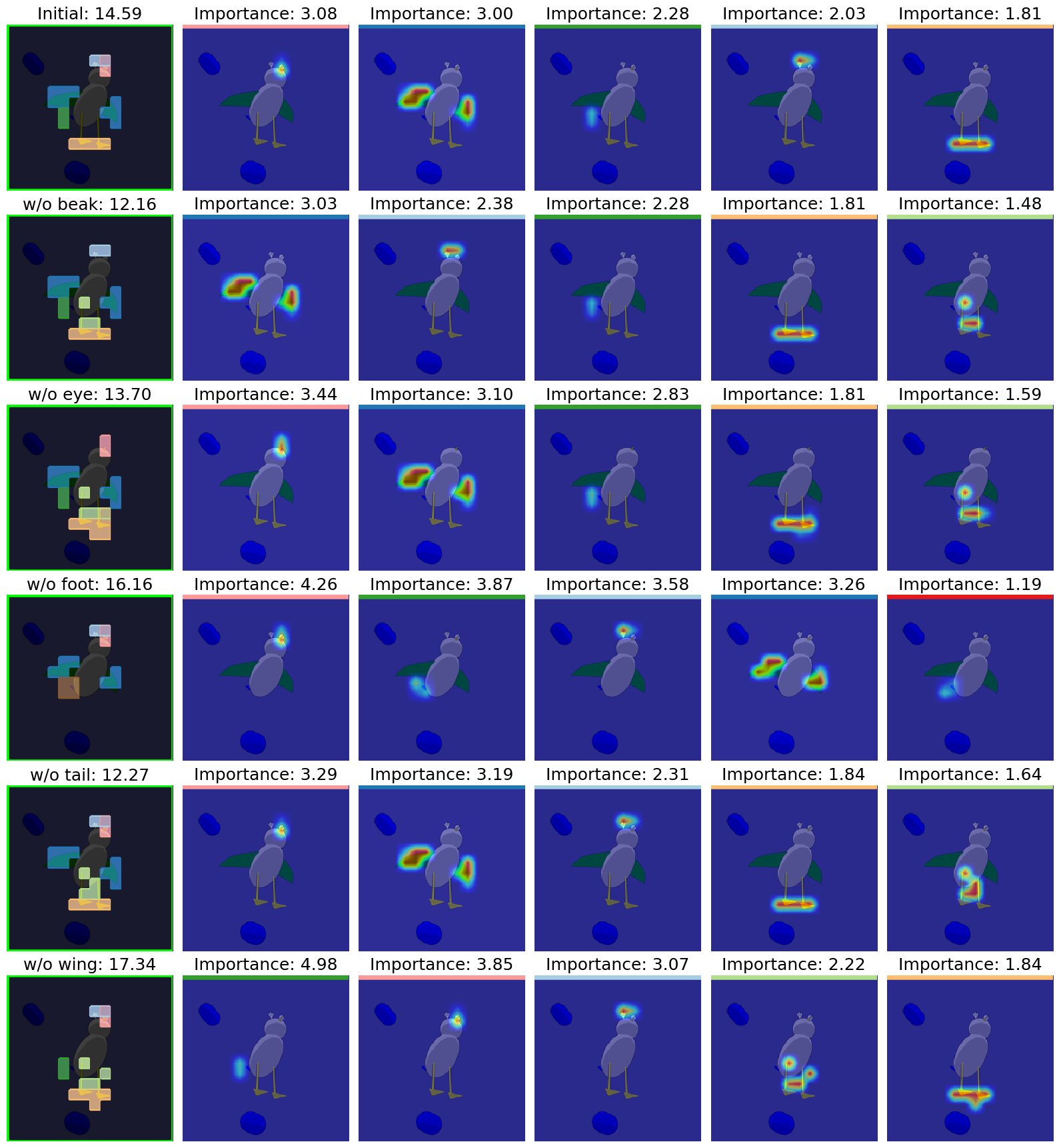}
\caption{Part deletion analysis on a sample from the FunnyBirds dataset. The first row represents the initial prediction on the non-corrupted sample. The following rows show the model's predictions along with the most important prototypes as different parts of the bird are removed. This figure allows to compare the importance attribution of each part with the change in score as this part is removed. }
\label{fig:sd_1}
\end{center}
\end{figure}

\begin{figure}[ht]
\begin{center}
\includegraphics[width=0.8\textwidth]{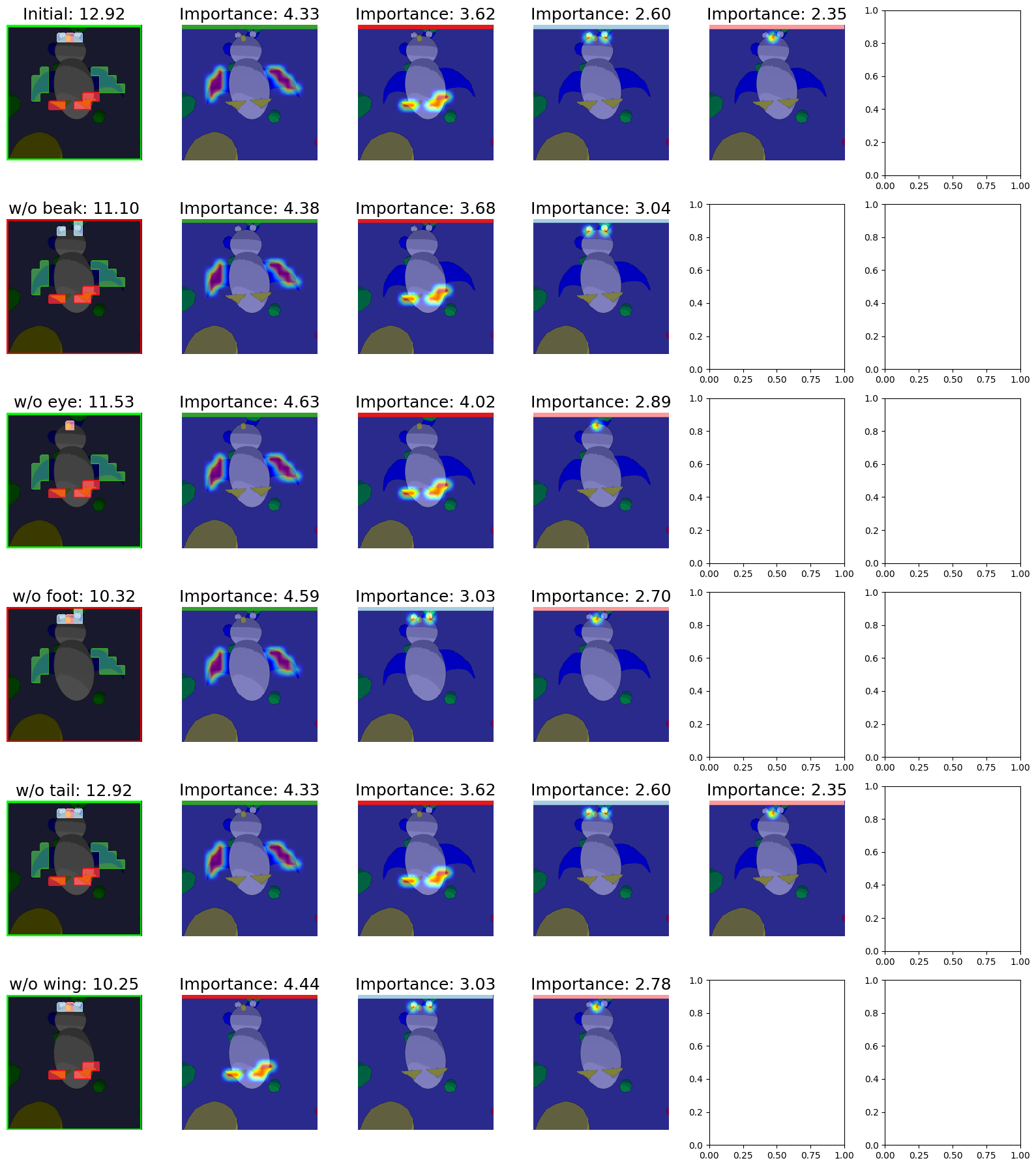}
\caption{Part deletion analysis on a sample from the FunnyBirds dataset. The first row represents the initial prediction on the non-corrupted sample. The following rows show the model's predictions along with the most important prototypes as different parts of the bird are removed. This figure allows to compare the importance attribution of each part with the change in score as this part is removed.}
\label{fig:sd_2}
\end{center}
\end{figure}

 The lowest metric for our approach is the single deletion (SD) metric. This metric evaluates whether the relevance attributed to each category: beak, eye, foot, tail and wing is correlated to their influence on the model's predictions. \cref{fig:sd_1} and~\cref{fig:sd_2} help illustrate how the model might be affected as different parts of the birds are removed. First, we observe that as the different parts are individually deleted, the corresponding prototype disappears reinforcing the strong spatial ability of the model. With this experiment, we can see that the local information encoded in the patch embeddings is directly related to the parts highlighted by the similarity. Regarding the single deletion metric, we observe that as a prototype is deleted, this prototype effectively disappears, but the importance of the other remaining prototypes increases. With this increase, the drop in score observed in the predictions with deleted parts cannot be directly related to the importance of the parts and the metric penalizes the model for this increase. However, this increase in the score of the prototypes might also help the model to be more robust as in most cases it is able to make a correct prediction and exploit the redundancy of the parts found in this specific dataset to make a correct prediction. 

 \begin{table}[]
\centering
  \caption{Explainability metrics evaluated on the FunnyBirds dataset. Top three metrics come from \cite{hesse2023funnybirds}, while the last two are adapted from \cite{huang2023evaluation}. }
  \label{tab:explainability_evaluation}
    \begin{tabular}{lccc}
        \toprule
    \multicolumn{1}{c}{Metric} & \multicolumn{3}{c}{Value} \\
 &  ST-ProtoPNet & PIP-NET & ProtoS-ViT (ours) \\  \midrule
        Completeness    & 70  & 62 & \textbf{92} \\
        Correctness     & 52  & 60 & \textbf{61}\\
        Contrastivity   & 64  & 1 & \textbf{99}\\
        Consistency & 55  & NA & \textbf{74} \\
        Stability & \textbf{99}  & NA & \textbf{99}\\  \bottomrule
\end{tabular}
\end{table}

The metrics from \cite{hesse2023funnybirds} have been calculated for ST-ProtoPNet, PIP-NET and ProtoS-ViT along with the adaptation of Consistency and Stability. They are presented in Table~\ref{tab:explainability_evaluation} in percentage rather than between zero and one to match the results in Figure~\ref{fig:radar_plot}. Contrastivity and Stability could not be calculated for PIP-NET as the explanation was a single patch not overlapping with any bird part, preventing us from running the analysis. Retaining local information is key for explainability. Indeed if the model shows an explanation which uses information not contained in the highlighted relevant patches, the model fails to provide transparent explanations. An indication that the local information is not retained is the contrastivity metric which is almost equal to zero.

 \clearpage
\section{Consistency and stability metrics adaptation}
\label{app:interpretability_metrics}
The consistency and stability metrics initially developed by Huang et al.~\cite{huang2023evaluation} were adapted to the FunnyBirds dataset. The aim was to allow a finer evaluation of the prototypes by taking advantage of the part-segmentation provided with this dataset. Both metrics are based for each image on the vector $o_\textbf{p}$. This vector is a binary vector indicating whether prototype $p_j$ is related to category $q\in Q$. There are five categories for the FunnyBirds dataset: beak, eye, foot, tail and wing. For each category we set the entry of the vector $o_\textbf{p}$ to one if an entry of the similarity map $\mathcal{M}_j$ weighted by the importance of the corresponding prototype $i_{j,k}$ is larger than 0.1 within the  binary segmentation mask corresponding to the given category $N_q$:
\begin{equation}
    o_{\textbf{p}_j}^q =\max \left \{i_{j,k} \left( \mathcal{M}_j \circ N_c \right) \right \} >0.1
\end{equation}
The consistency and stability scores are then evaluated using the same formula as~\cite{huang2023evaluation} with our modified vector $o_\textbf{p}$. However as the initial paper considers prototypical models where prototypes only belong to one class, we repeat the operations across all classes.  Only the prototype that appears in the prediction for the considered class is included, and the result is averaged across all classes. 
\section{Impact of backbone}
\label{app:backbone}

This section presents results aiming to better understand the choice of the backbone on model performance, for both ProtoS-Vit and PIP-Net. In order to understand the impact of the backbone on PIP-Net, this architecture was retrained with DINO ViT-B/14 and evaluated on the FunnyBirds dataset. This change allow to compare our architecture with PIP-Net with the same backbone. Results presented in Table~\ref{table:backbone_study} show that while the model with a trainable backbone performs better in terms of explanation quality across different metrics, it still suffers from the contrastivity metric equal to zero, meaning it does not retain local information, which is key for explainability.

\begin{figure}[H]
    \centering
    \includegraphics[width=0.99\linewidth]{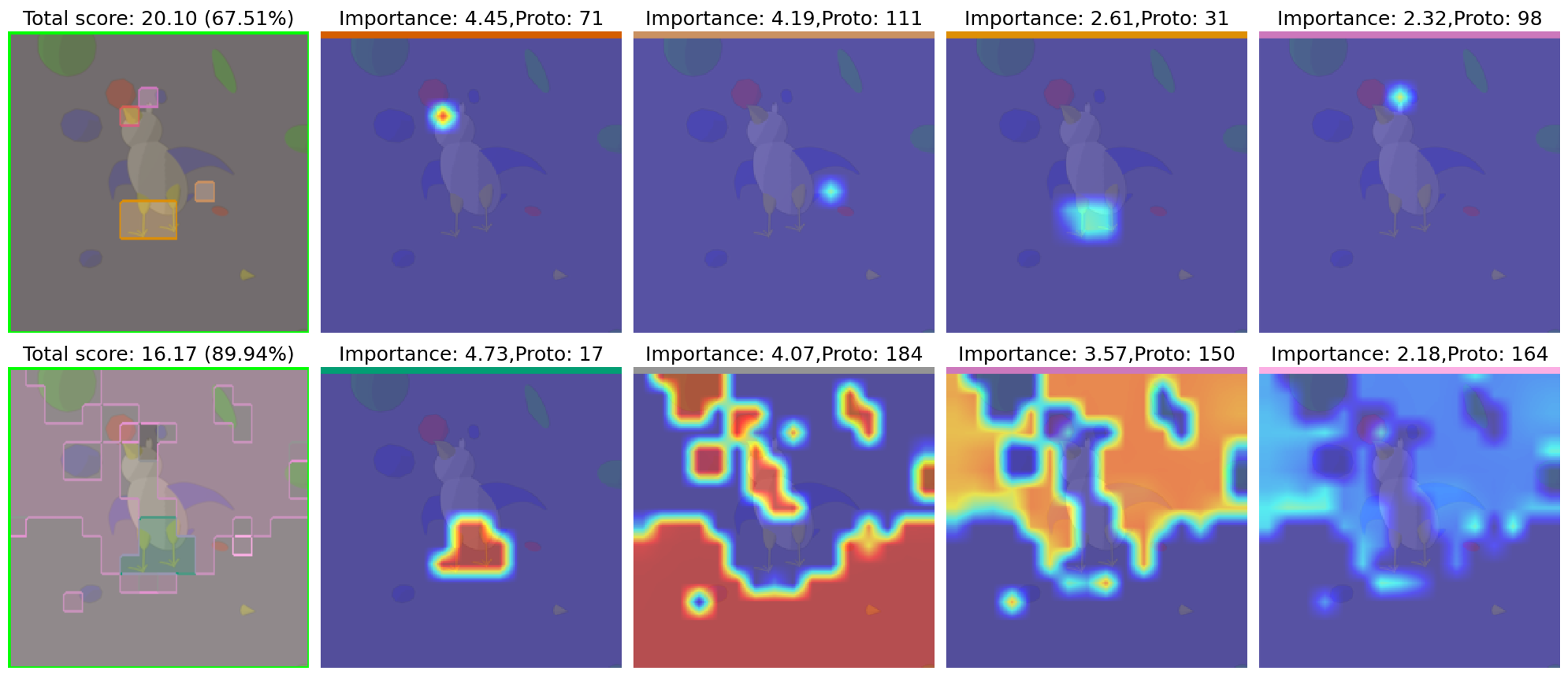}
    \caption{Score sheet for predictions on two random samples from the FunnyBirds dataset. Top row: frozen backbone, Bottom row: trainable backbone.}
    \label{fig:score_sheet_trained}
\end{figure}

Table~\ref{table:backbone_study} also shows the explanation metrics comparing ProtoS-Vit when freezing or training the backbone. Overall, we observe that training the backbone greatly reduces the quality of the explanation provided by the model especially the consistency of the metric. An example of a score sheet obtained when the backbone is trained is shown in
Figure~\ref{fig:score_sheet_trained}.

\begin{table}[h!]
\caption{Comparison between ProtoS-Vit and PIP-NET on explainability metrics evaluated on the FunnyBirds dataset. BI stands for background independence. \textbf{Bold} indicates the best score for the given metric.}
\label{table:backbone_study}
\begin{tabularx}{\textwidth}{XYYY}
\toprule
Architecture            & \multicolumn{2}{c}{ProtoS-ViT}                         & PIP-Net                     \\ \midrule
Backbone                & DINO ViT-B/14\newline (Freeze) & DINOv2 ViT-B/14 \newline(Trainable) & DINO ViT-B/14 (Trainable) \\
Accuracy                & 0.96                     & 0.95                        & \textbf{0.99}                        \\
CSDC                    & \textbf{0.94}                     & 0.92                        & 0.60                        \\
PC                      & \textbf{0.96}                     & 0.90                        & 0.43                        \\
DC                      & \textbf{0.95}                     & 0.87                        & 0.39                        \\
Distractability         & 0.89                     & 0.84                        & \textbf{0.93}                       \\
BI & 0.99                     & \textbf{1.00}                        & \textbf{1.00}                        \\
SD                      & 0.63                     & \textbf{0.76}                        & 0.70                        \\
TS                      & \textbf{0.99}                     & 0.95                        & \textcolor{red}{\textbf{0.00}}                        \\
Consistency             & 0.70                     & 0.57                        & \textbf{1.00}                        \\
Stability               & 0.99                     & 0.97                        & \textbf{1.00}                        \\ \bottomrule
\end{tabularx}
\end{table}

\clearpage
\section{Evaluation of prototypes quality and semantical consistency}

\subsection{Classification head correlation}
\label{app:correlation_matrix}
\begin{figure}[H]
\centering
\includegraphics[width=1\textwidth]{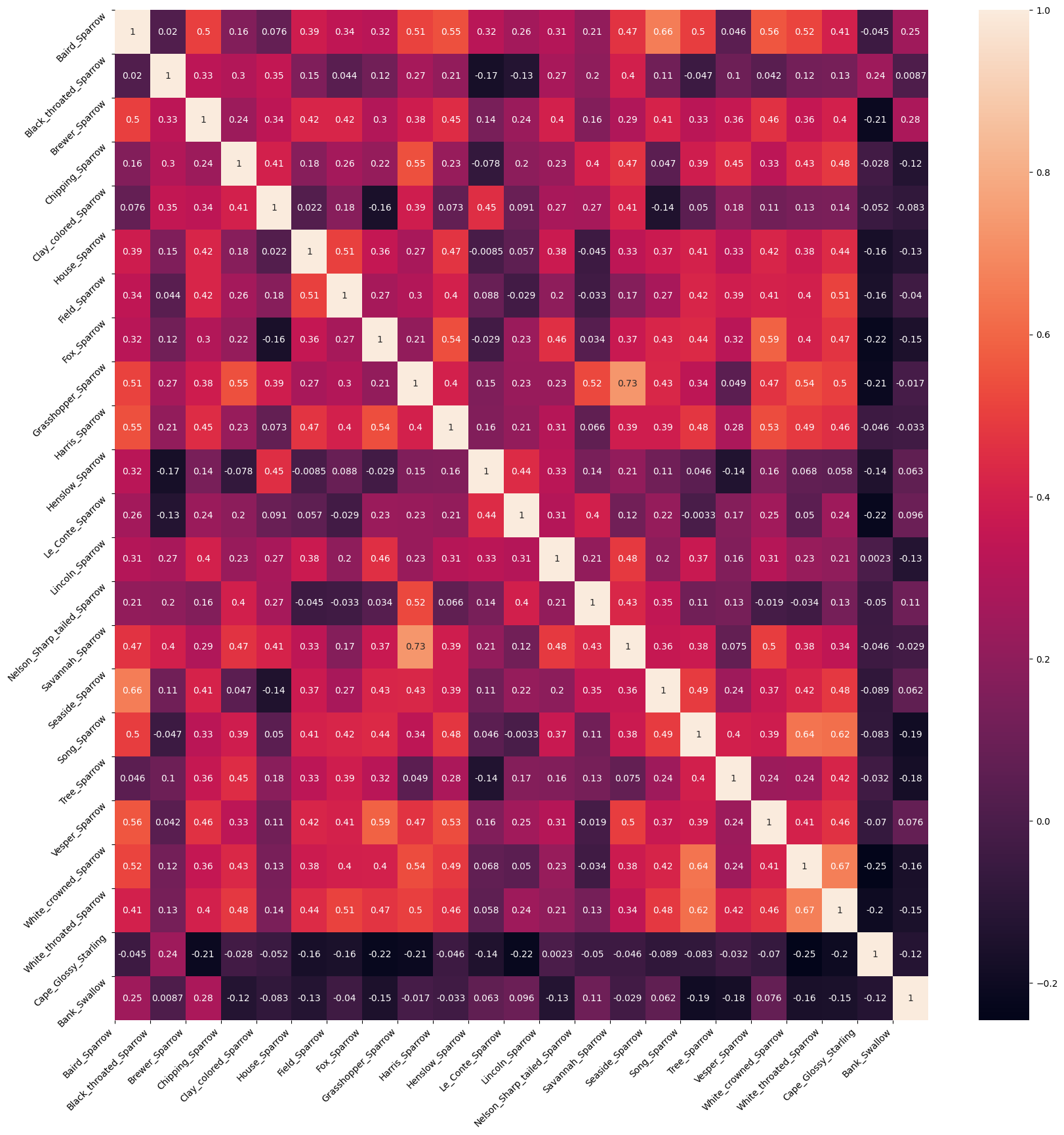}

\caption{Classification head correlation matrix for classes 112 to 135 of the CUB dataset}
\label{fig:correlation_matrix_1}
\end{figure}

\begin{figure}[H]
\begin{center}
\includegraphics[width=1\textwidth]{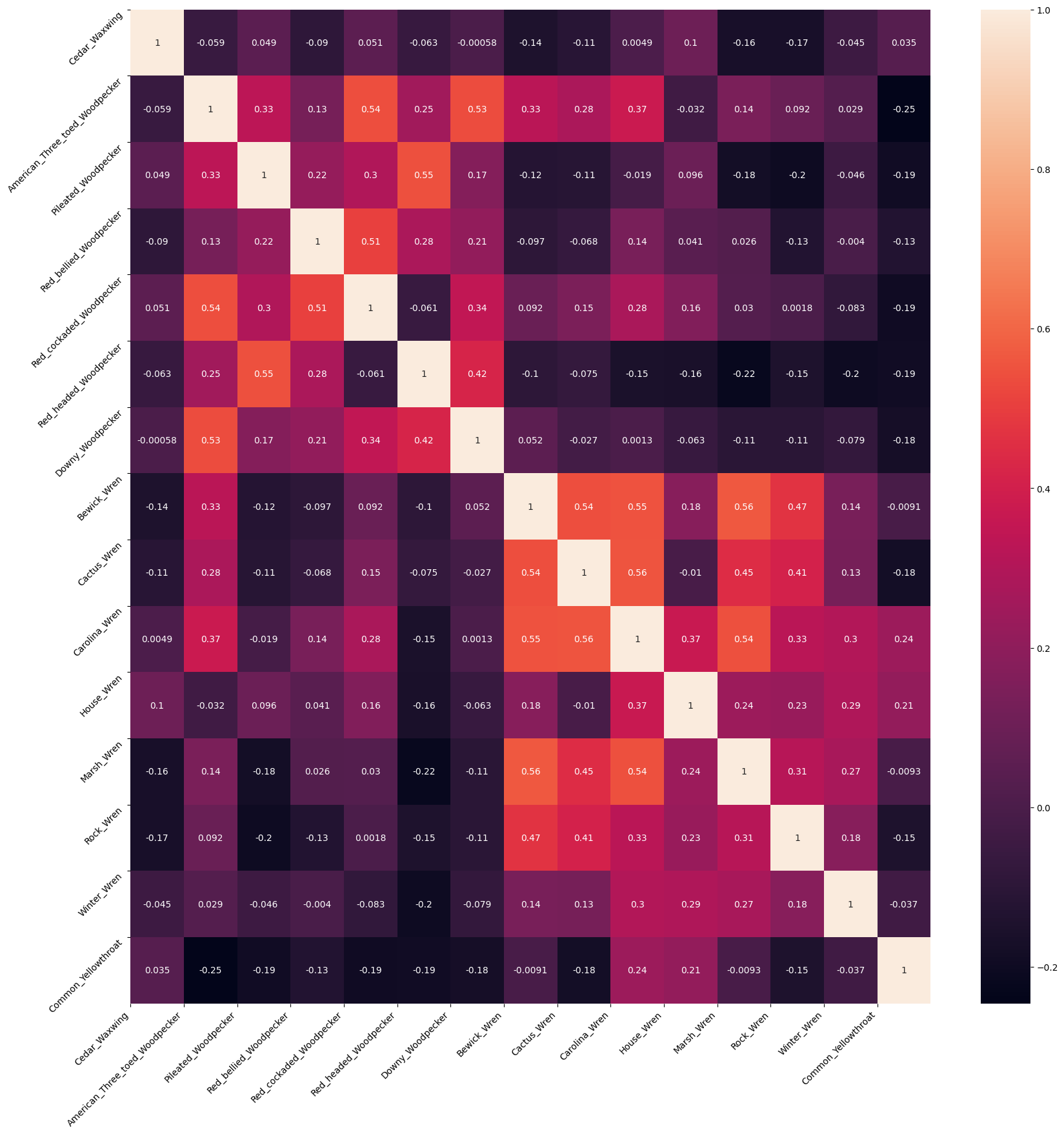}

\caption{Classification head correlation matrix for classes 112 to 135 of the CUB dataset}
\label{fig:correlation_matrix_2}
\end{center}
\end{figure}

To qualitatively evaluate how the prototypes are reused across classes, we also looked at the correlation of the weights from the classification head. These weights assign prototypes to the different classes. Analyses of the correlation across classes of the CUB dataset show that subspecies from a common species have a high correlation across their corresponding vector in the classification weights as measured using the Pearson correlation coefficient. Figure~\ref{fig:correlation_matrix_1} shows a strong correlation across sparrow subspecies while  Figure~\ref{fig:correlation_matrix_2} shows the same level of correlation across both woodpecker and wren. Overall, this analysis shows that prototypes are shared across subspecies effectively sharing prototypes across similar classes.
\subsection{User Study}
\label{app:User-Studies}
In addition to the five quantitative metrics used to assess the quality of the explanations provided by the designed architecture, an additional user-study was carried to better understand the consistency of the prototypes with respect to concepts human would associate together as well as their relevance towards the classifications tasks. The user-study rely on a random selection for each prototype of 100 samples where this prototype is playing a role toward the model's prediction. This user-study was carried on the Funny-Birds dataset. Indeed this dataset was designed so that the discriminative portion of each image is well defined by meta-features: the eyes, beak, wings, legs and tail. The samples used for the two user-studies can be found in Supplementary Materials~\cite{zenodo}.

\begin{figure}[H]
\begin{center}
\includegraphics[width=1\textwidth]{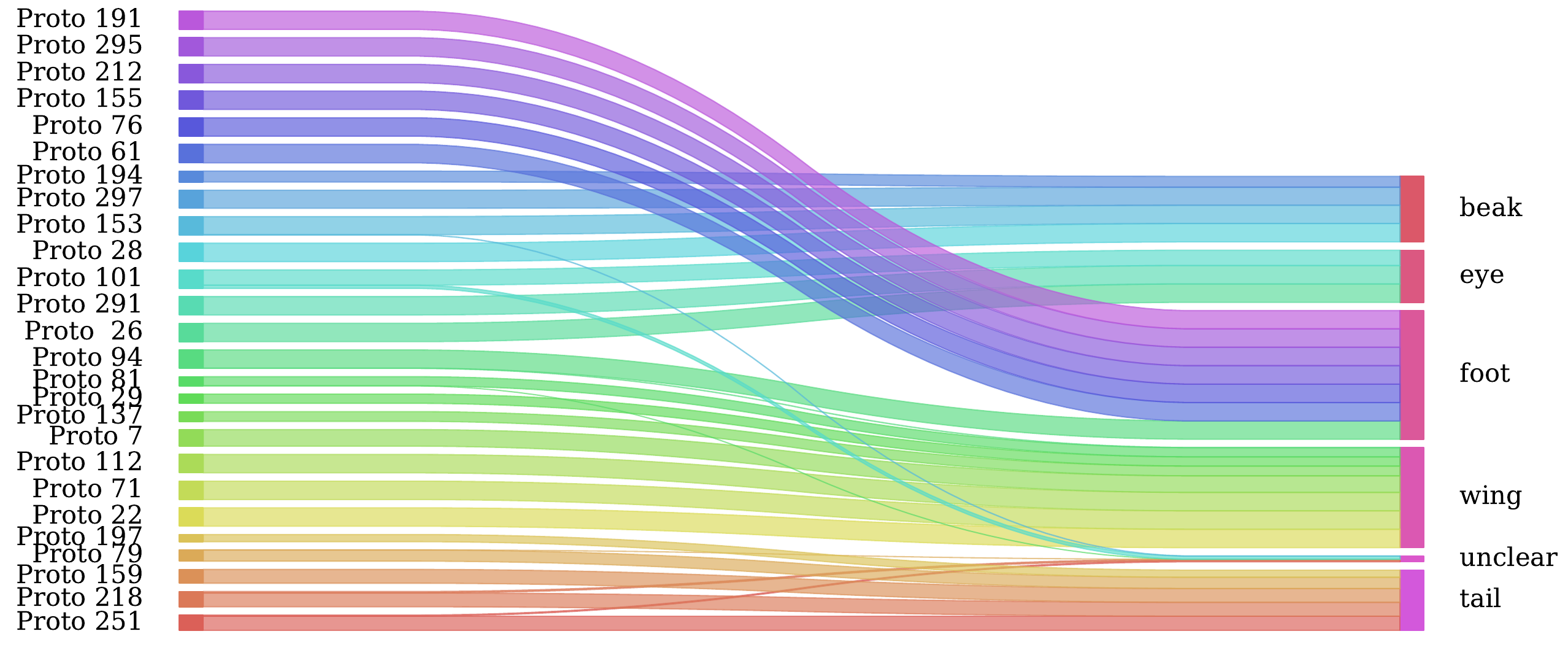}
\caption{User-Study performed on the Funny-Birds datasets. The user was asked to assess the prototype visual consistency of the prototype (whether a prototype is associated to a specific bird part). 100 samples were visualised per prototype, when available (some prototypes were not present 100 times in the test set.)}
\label{fig:us_1}
\end{center}
\end{figure}
The authors of the FunnyBirds dataset manually designed 5 bird parts: eyes (3 types), beak (4 types), wings (6 types), legs (4 types) and tail (9 types) to construct the 50 classes. As depicted in Figure~\ref{fig:us_1}, the learned prototypes were attributed consistently to the same parts with the following number of prototypes per part: eye (3 prototypes), beak (4 prototypes), wings (7 prototypes), legs (7 prototypes) and tail (5 prototypes). The consistency of the prototype was then evaluated by counting how many times each prototype highlighted the same region of the bird. It was found that 21 prototypes scored 100\%, 2 prototypes scored 99\%, 1 scored 93\%, 1 scored 90\%, and one scored 83\%. For the eyes and beaks parts, the number of learned prototypes match exactly the number of bird part types. Each prototype can therefore be directly attributed to a specific part, e.g. prototype \#101 relates to "eye0", prototype \#297 relates to "beak1". This user study allows to confirms the \textit{meaningfulness} of the prototypes derived from the proposed architecture, as well as \textit{compactness} of the explanations allowing direct comparison with how a human would approach the classification task. Full results for this study can be found in Supplementary Materials~\cite{zenodo}.

\end{document}